\documentclass{article}
\usepackage{amsmath,epsfig}
\usepackage[preprint]{spconfa4}
\usepackage{xcolor}
\usepackage{cite}
\usepackage{multirow}
\usepackage{amssymb}

\let\OLDthebibliography\thebibliography
\renewcommand\thebibliography[1]{
  \OLDthebibliography{#1}
  \setlength{\parskip}{0pt}
  \setlength{\itemsep}{0pt plus 0.3ex}
}

\begin{document}\sloppy

\def\x{{\mathbf x}}
\def\L{{\cal L}}

\title{Question-Driven Graph Fusion Network For Visual Question Answering}
%
\name{Yuxi Qian, Yuncong Hu, Ruonan Wang, Fangxiang Feng, Xiaojie Wang\textsuperscript{*}}
\address{Beijing University of Posts and Telecommunications \\
\{qianyuxi, hycthan, wangrn, fxfeng, xjwang\}@bupt.edu.cn}

\maketitle

\begin{abstract}
Existing Visual Question Answering (VQA) models have explored various visual relationships between objects in the image to answer complex questions, which inevitably introduces irrelevant information brought by inaccurate object detection and text grounding. To address the problem, we propose a \textbf{Question-Driven Graph Fusion Network (QD-GFN)}. It first models semantic, spatial, and implicit visual relations in images by three graph attention networks, then question information is utilized to guide the aggregation process of the three graphs, further, our QD-GFN adopts an object filtering mechanism to remove question-irrelevant objects contained in the image. Experiment results demonstrate that our QD-GFN outperforms the prior state-of-the-art on both VQA 2.0 and VQA-CP v2 datasets. Further analysis shows that both the novel graph aggregation method and object filtering mechanism play a significant role in improving the performance of the model.
\end{abstract}
\section{Introduction}
\renewcommand{\thefootnote}{}
\footnote{
*Xiaojie Wang is the corresponding author.
}
\label{sec:intro}
\noindent
Visual Question Answering \cite{Agrawal2015VQAVQ} is one of the most challenging multi-modal tasks. Given an image and a natural language question, the task is to correctly answer the question by making use of both visual and textual information. During the last few years, Visual Question Answering (VQA) has attracted rapidly growing attention. Since most questions not only focus on the objects in the image but also need to be answered by combining with the relationship between them. Visual relationship, therefore, plays a crucial role in VQA. Considering this, \cite{Cadne2019MURELMR, Li2019RelationAwareGA, Hu2019LanguageConditionedGN} adopt graph neural networks to comprehensively capture inter-object relations contained in images.

\begin{figure}
    \begin{center}
    \includegraphics[width=1.0\linewidth]{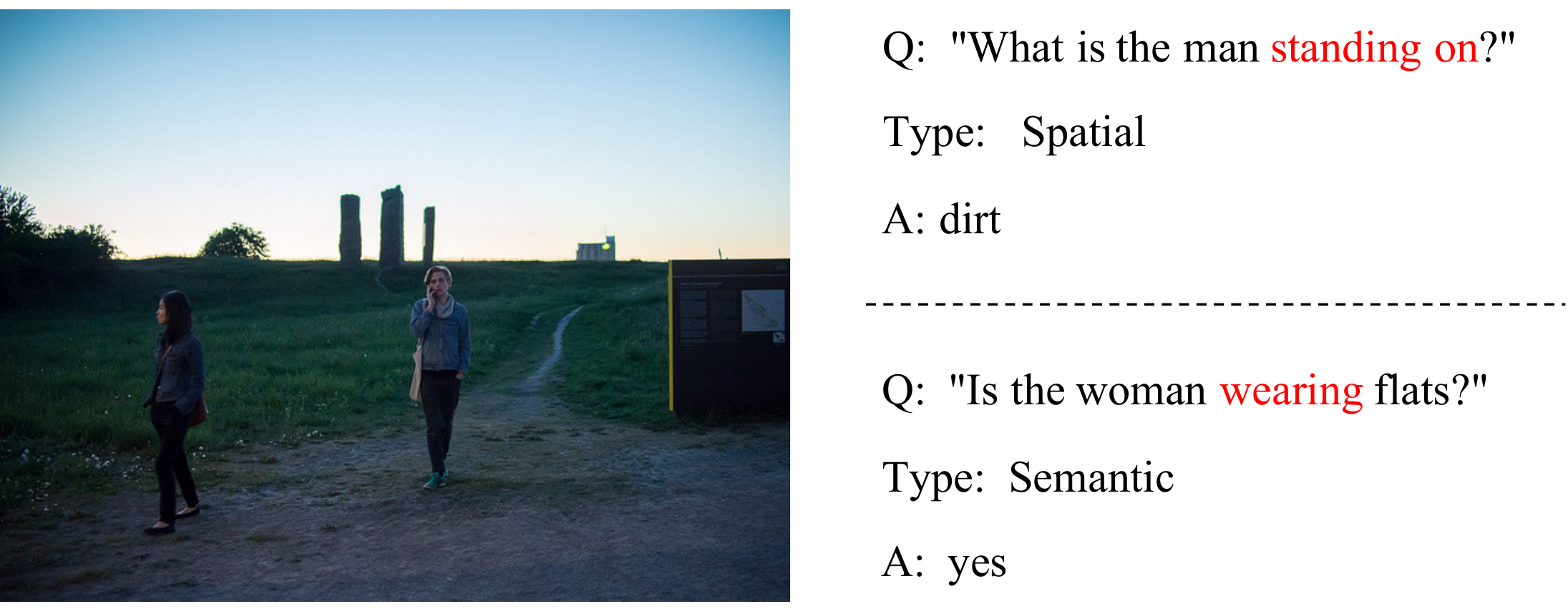}
    \end{center}
    \caption{A case for illustration, different questions focus on different objects and types of relationship in the image}
    \label{fig:case}
\end{figure}

Although the above models have achieved better performances by exploring various relation features, a large amount of irrelevant information is also introduced, which affects the final performances of the above models. As shown in Figure~\ref{fig:case}, the first question for the image can be answered just by considering spatial relations while the other one only concentrates on semantic relations. Moreover, it is obvious that both the two questions only connect with a few objects. To coordinate different relationships properly and reduce the negative impact of the noises caused by redundant objects, we propose a novel model for VQA, named \textbf{Question-Driven Graph Fusion Network(QD-GFN)}.

In general, in the first stage, to fully cover the rich relationship information in the image, inspired by \cite{Li2019RelationAwareGA}, we use multiple different graph attention networks to capture different types of visual relationship. Further, to accomplish question-guided graph fusion, a novel method is designed based on a cross-attention mechanism to measure correlation degree between question and each type of relationship, then these relationships between objects are updated according to the above correlation degree. While for object filtering, we propose an object priority coefficient, and use such coefficient to remove objects with low importance.

In principle, our work provides a new perspective for exploring VQA tasks. The contributions of our work can be concluded as follows:
\begin{itemize}
\item We propose a novel question-guided graph fusion module to better coordinate different type of relationships.
\item We propose an object filtering mechanism to reduce the interference caused by irrelevant objects.
\item Our method achieves superior performance on the VQA 2.0 dataset \cite{Goyal2017MakingTV} and the more challenging VQA-CP v2 dataset \cite{Agrawal2018DontJA}.
\end{itemize}
\section{Related Work}
\subsection{Visual Question Answering}
\noindent
Conventional framework for VQA systems \cite{Agrawal2015VQAVQ} consists of an image encoder, a question encoder, multimodal fusion, and an answer predictor. In the past few years, \cite{Yu2017MultimodalFB, Benyounes2017MUTANMT} have explored new multimodal feature fusion strategies to better combine visual and textual information in high-dimensional space. Meanwhile, to better understand the visual contents of images and the semantics of questions, \cite{Lu2016HierarchicalQC, Yu2017MultimodalFB} adopt a coarse co-attention. However, above coarse co-attention neglects the relationship between each image region and each question word. To fully excavate the above information, BAN \cite{Kim2018BilinearAN} establish dense interactions between each image region and each question word. Recently, with the proposal of Transformer \cite{Vaswani2017AttentionIA}, \cite{Gao2019DynamicFW, Gao2019MultiModalityLI} utilize multi-head attention to excavate the fine-grained implicit relationship in both inter- and intra-modality. To improve the interpretability of model, \cite{Li2019RelationAwareGA, Cadne2019MURELMR} directly encode explicit relation, such as semantic relation and spatial relation between objects, into image representation.

Our work is complementary to above studies. Based on \cite{Li2019RelationAwareGA}, we further propose a question-guided graph fusion module to better aggregate the information contained in different types of graphs and introduce an object filtering mechanism to reduce interference caused by irrelevant information.
\subsection{Visual Relationship}
\noindent
 Early work treat Visual Relationship Detection(VRD) \cite{Divvala2009AnES, Galleguillos2008ObjectCU} as a post-processing step for object detection, the detected objects are re-scored by considering object relations contained in the image. Since utilizing visual relationships between objects is of great benefit to many computer vision and multimodal tasks, \cite{Lu2016VisualRD} attempts to capture a variety of relationships between objects more sufficiently. Recently, with the propose of Scene Graph Generation (SGG) task, \cite{Zellers2018NeuralMS, Tang2020UnbiasedSG} reduce the interference caused by bias and further improves the quality of relationship extraction.
\begin{figure}
    \begin{center}
    \includegraphics[width=1.0\linewidth]{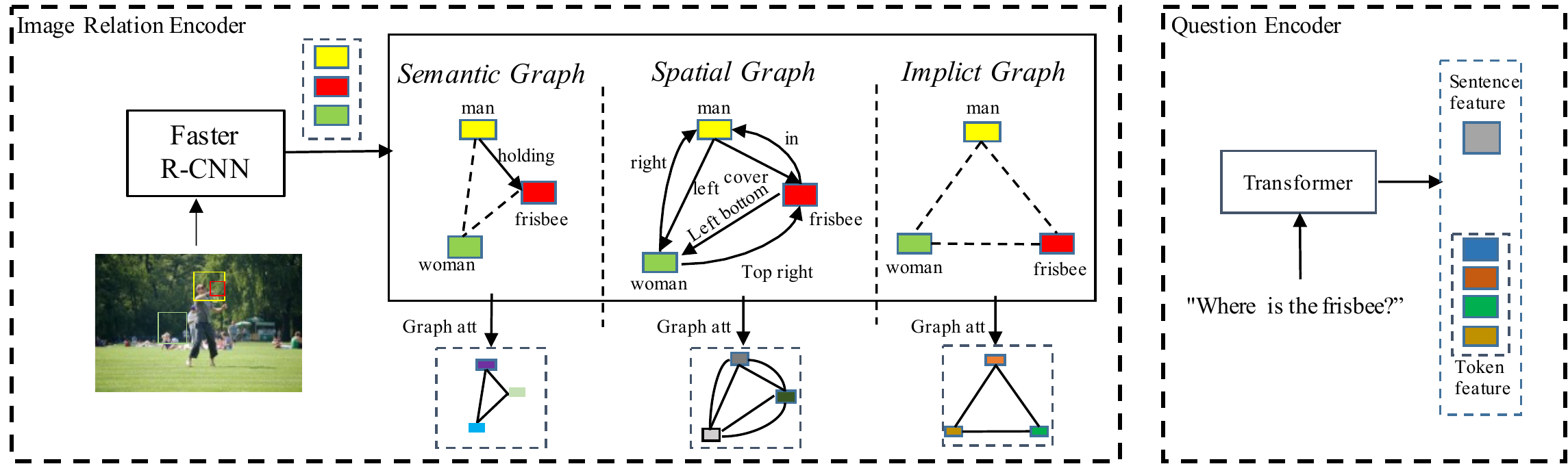}
    \end{center}
    \caption{Structure of Image Relation Encoder and Question Encoder, the image relation encoder consists of three graph attention networks: semantic graph, spatial graph and implicit graph}
    \label{fig:encoder}
\end{figure}
\begin{figure*}[t]
    \begin{center}
    \includegraphics[width=0.8\linewidth]{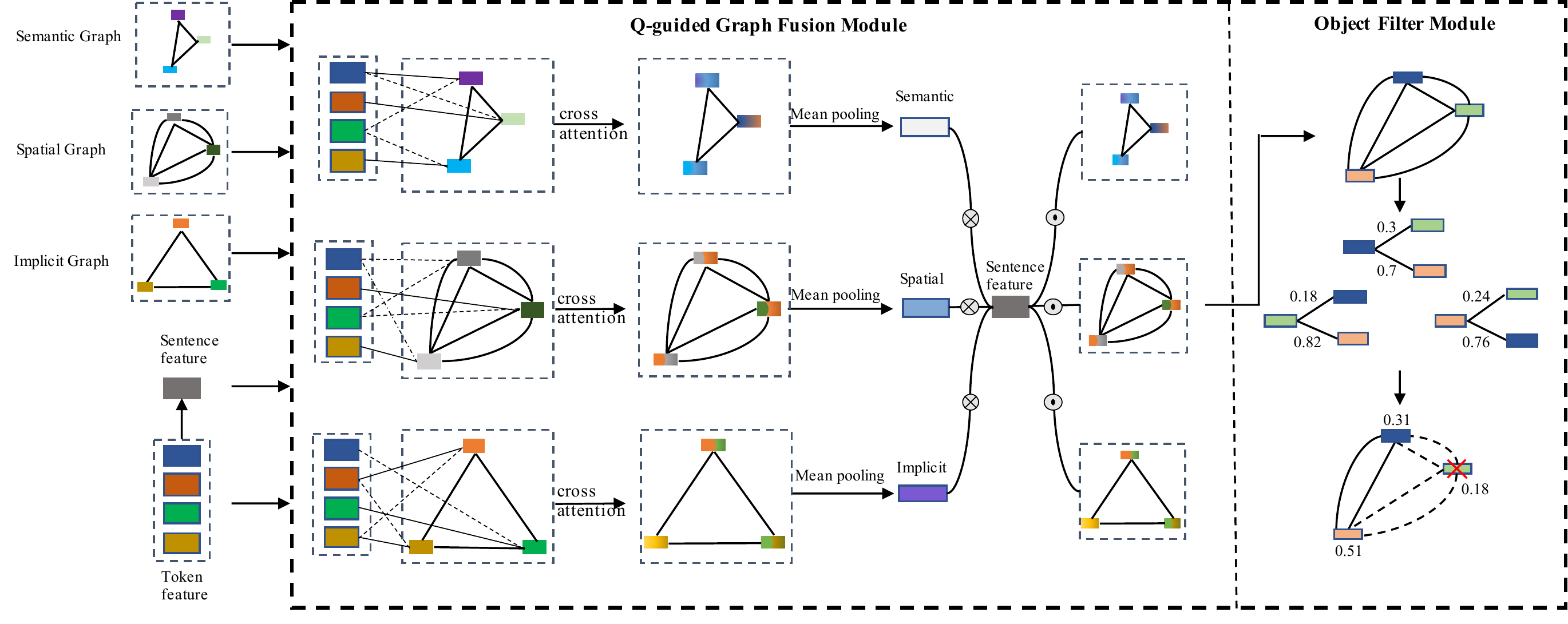}
    \end{center}
    \caption{Detailed illustration of our proposed Question-guided Graph Fusion Module and Objects Filter Module. Relation-aware visual features and context-aware textual features are obtained through Image relation Encoder and Question Encoder respectively. Question-guided graph fusion module is employed to re-weight different types of relationships based on the question, and aggregate multiple graph networks. Object Filter will remove objects which are less relevant to the question.}
    \label{fig:model}
\end{figure*}

\section{Method}
\noindent
Given a question $q$ and its corresponding image $I$ , the goal of VQA is to predict an answer $a \in A$ that best matches $q$. This task is defined as a classification problem \cite{Agrawal2015VQAVQ}:
\begin{equation}
    \hat{a} =\mathop{argmax}\limits_{{a} \in {A}}P_{\theta}(a|I,q),
\end{equation}
where $\theta$ is the parameters of the model.
The detailed architecture of our QD-GFN is displayed in Figure~\ref{fig:encoder} and Figure~\ref{fig:model}. Relation-aware visual features and textual features are obtained from Image Relation Encoder and Question Encoder respectively. Then, Question-guided Graph Fusion Module adaptively aggregates the rich information contained in the image according to the question and Object Filtering Module further eliminates the interference caused by irrelevant objects.
\subsection{Question Encoder and Image Relation Encoder}
\noindent
Question Encoder is mainly implemented on a Transformer model \cite{Vaswani2017AttentionIA}, question will be padded to a maximum length of 20 and be encoded by Transformer with random initialization, denoted as $T=\{t_1, t_2, ...t_n\}$, where $t_i \in \mathbb{R}^{768}$. For the Image Relation Encoder, Faster R-CNN \cite{Ren2015FasterRT} is first employed to recognise objects $O=(o_1,o_2...o_m)$ in $I$, where each object $o_i$ is composed of a visual feature $ v_i \in \mathbb{R}^{2048}$ and a bounding-box feature $ b_i \in \mathbb{R}^{4}$. Inspired by \cite{Li2019RelationAwareGA}, we adopt three question-adaptive graph attention networks \cite{Velickovic2018GraphAN} to encode semantic relations, spatial relations and implicit relations into different image representations. For each $v_i$ in the graph, we introduce multi-head attention mechanism to calculate its correlation coefficient with other objects $v_j$ as $\alpha_{ij}$, and select objects with top K coefficients as its adjacency point sets $N_i = topk(\alpha_{ij})$, therefore, for each head, $\alpha^{h}_{ij}$ can be denoted as, where $W^{q}_{V},W^{k}_{V} \in \mathbb{R}^{768 \times 768}$:
\begin{equation}
  \alpha^{h}_{ij} = (W^{q}_{V} v_i)^{T} \cdot W^{k}_{V}v_j
\end{equation}
\textbf{Implicit Graph}: Due to there is no prior information introduced into the implicit graph $G_{imp}$, edges in implicit graph can be directly illustrated as:
\begin{equation}
    e^{h}_{ij} = \frac {\exp (\alpha^{h}_{ij})}{\sum_{j \in N_i}\exp(\alpha^{h}_{ij})}
\end{equation}
\textbf{Explicit Graph}: Since edges in the semantic graph $G_{sem}$ and the spatial graph $G_{spa}$  contain label information, we modified multi-head attention mechanism to be sensitive to explicit relation labels:
\begin{equation}
    e^{h}_{ij} = \frac {\exp (\alpha^{h}_{ij} + c_{lab(i,j)})}{\sum_{j \in N_i}\exp(\alpha^{h}_{ij} + c_{lab(i,j)})} 
\end{equation}
where $c_{lab(i,j)}$  represents the explicit relation labels. The update process for the relational graphs can be illustrated as following, where $W_V^h \in \mathbb{R}^{768 \times 768} $:
\begin{equation}
    v_{i}^{'}={||}_{h=1}^{H}\sigma(\sum_{j \in N^{h}_{i}} e^{h}_{ij} \cdot W_V^h v_j)
\end{equation}
\subsection{Question-guided Graph Fusion Module}
\noindent
Question-guided Graph Fusion Module(GFM) aims to adaptively aggregate different types of relations and visual features according to their relevance to question. Specially, after obtaining the relation-aware visual features, we adopt visual-guided question attention $\alpha^{v->q}$ to inject information from the question into visual features:
\begin{equation}
     \alpha^{v->q}_{ij} = \frac {\exp (W^{q}_{V} v^{'}_i \cdot (W^{k}_{Q}t_j)^{T})}{\sum_{j=1}^{L}\exp(W^{q}_{V}v^{'}_i \cdot (W^{k}_{Q}t_j)^{T})}
\end{equation}
where L denote the length of questions, $W^{k}_{Q}, W^{q}_{V} \in \mathbb{R}^{768 \times 768} $. As we adopt multi-head attention, the update of visual features can be depicted as:
\begin{equation}
    v_i^*=v_{i}^{'} + {||}_{h=1}^{H}\sigma(\sum_{j=1}^{L} \alpha^{h, v->q}_{ij} \cdot W_{Q}^{h,v} t_{j})
\end{equation}
where $W_{Q}^{h,v} \in {\mathbb{R}^{768 \times 768}}$. To measure the correlation between question and each relation graph, mean pooling is preformed on question feature and each relation-aware graph feature to attain global representation $Q$ and $G_{k}$, where $k$ represent the k-th graph, then we calculate the cos similarity between  $Q$ and $G_{k}$, denoted as $\beta_{k}$:
\begin{equation}
    \beta_{k} = \frac{cos(Q, G_k)}{\sum_{i=1}^{3}cos(Q, G_i)}
\end{equation}
where $G_{k}$ represents the k-th relation graph in graph sets.
Finally, $\beta_{j}$ are utilized to guide the fusion process of above graphs:
\begin{gather}
    \hat{v}_{i} = \sum_{k=1}^{K} \beta_{k} * v_{ik}^*, \\
    \hat{\alpha}_{ij} = \sum_{k=1}^{K} \beta_{k} * \alpha_{ijk}
\end{gather}
and we denote the graph obtained after aggregation as $\hat{G}$.
\subsection{Objects Filtering Module}
\noindent
To further reduce the interference caused by irrelevant information, Objects Filter Module (OF) is deployed to remove objects that are less important. Therefore, we introduce the object priority coefficient which is based on attention to measure the importance of each object $o_i$ in $O$.\\
\textbf{Object Priority Coefficient($\gamma$)} is proposed to precisely measure the importance of each object. $\hat{G}$ is regarded as a fully-connected graph, we calculate the correlation coefficient between two objects based on attention:
\begin{equation}
    \hat{e}_{ij} =   \frac {\exp ((W^{q}_{V} \hat{v}_i)^{T} \cdot W^{k}_{V} \hat{v}_j+ \hat{\alpha}_{ij} )}{\sum\exp(((W^{q}_{V}) \hat{v}_i)^{T}\cdot W^{k}_{V} \hat{v}_j + \hat{\alpha}_{ij})}
\end{equation}
 where $W^{q}_{V},W^{k}_{V} \in \mathbb{R}^{768 \times 768}$. For the i-th object,its priority coefficient $\gamma_{i}$ can be calculated as:
\begin{equation}
    \gamma_{i} = \frac{(\sum_{j=1}^{m} \hat{e}_{ji})^2}{\sum_{i=1}^{m}(\sum_{j=1}^{m} \hat{e}_{ji})^2}
\end{equation}

Then we select the top P objects with the largest $\gamma$ as the effective objects set, which denoted as $EO$ and update the relation between objects in $EO$:
\begin{equation}
    \Tilde{e}_{ij} =  \frac {\exp ((W^{q}_{V} \hat{v}_i)^{T} \cdot W^{k}_{V} \hat{v}_j )}{\sum_{i,j \in EO}\exp(((W^{q}_{V}) \hat{v}_i)^{T}\cdot W^{k}_{V} \hat{v}_j)}
\end{equation}

Based on the updated relation $\Tilde{e}$, we again calculate the $\gamma_{i}$ for the objects remained in $EO$, and perform the final aggregation as follows:
\begin{equation}
    V^{*} = \sum_{i \in EO} \gamma_{i} * \hat{v}_i
\end{equation}
\subsection{Answer Prediction}
\noindent
After obtaining the final visual features $V^{*}$, we fuse question information $q$ with $V^{*}$  to attain a joint representation $J$:
\begin{equation}
    J = q * V^*
\end{equation}
For Answer Predictor, we adopt a two-layer MLP as the classifier, with $J$ as the input. Binary cross entropy is utilized as the loss function.
\section{Experiment}
\noindent
We conduct experiments to evaluate the performance of our QD-GFN on the VQA 2.0\cite{Goyal2017MakingTV} and VQA-CP v2\cite{Agrawal2018DontJA} datasets. Meanwhile, we perform extensive ablation studies to explore the potential factors which may affect the final performance of the model.
\begin{table}
    \centering
    \resizebox{\linewidth}{!}{
    \begin{tabular}{l|c|c|c|c|c|c}
    \hline
    \multirow{2}{*}{Method} & Validation & \multicolumn{4}{c|}{Test-dev} & Test-std \\
    \cline{2-7} & All  & Y/N & Num & Other & All  &All \\
    \hline
    BUTD \cite{Anderson2018BottomUpAT}      & 63.37  & 81.82  & 44.21  & 56.05 & 65.32 & 65.67\\
    BAN + Counter \cite{Kim2018BilinearAN}     & 66.04   &85.42  & 54.04 & 60.52 & 70.04 & 70.35 \\
    MuREL\cite{Cadne2019MURELMR}    & 65.14   & 84.77 & 49.84 & 57.85 & 68.03 & 68.41 \\
    DFAF\cite{Gao2019DynamicFW}     & 66.66 & 86.09 & 53.32 & 60.49 & 70.22 & 70.34 \\
    MLIN\cite{Gao2019MultiModalityLI}     & 66.53   & 85.96  & 52.93 & 60.40 & 70.18  & 70.28\\
    ReGAT\cite{Li2019RelationAwareGA} & 67.18 & 86.08 & \textbf{54.42} & 60.33 & 70.27 & 70.58\\
    \hline
    Ours     & \textbf{67.61} & \textbf{86.45}  & 54.41  & \textbf{60.52} & \textbf{70.51} & \textbf{70.71} \\
    \hline
    \end{tabular}}
    \centering
	\caption{Model performance on VQA 2.0 benchmark.} 
	\label{tab:vqaval}
\end{table}
\begin{table}
    \centering
    \begin{tabular}{l|c|c|c|c}
    \hline
    Model & All & Y/N & Num & Other \\
    \hline
    RAMEN\cite{Shrestha2019AnswerTA}  & 39.21 & - & - & - \\
    BAN\cite{Kim2018BilinearAN} & 39.31 & - & - & -\\
    MuREL\cite{Cadne2019MURELMR} & 39.54 & 42.85 & 13.17 & 45.04  \\
    ReGAT\cite{Li2019RelationAwareGA} & 40.42 & - & - & - \\
    \hline
    Ours & \textbf{43.36} & \textbf{44.56} & \textbf{15.49} & \textbf{50.39} \\
    \hline
    \end{tabular}
    \vspace{0.1cm}\caption{Model performance comparison on the VQA-CP v2 dataset.}
    \label{tab:vqacp}
\end{table}

\subsection{Implementation Details}
\noindent
Each question is tokenized and padded with 0 to a maximum length of 20. We choose a 3-layer transformer encoder with 768 hidden size and 12 attention heads as the question encoder. For image relation encoder, we first extract pre-trained object features with bounding boxes from Faster R-CNN and then map them to the same dimension as the textual feature. We employ multi-head attention with 12 heads for all graph attention networks and set the number of adjacent points for each node to 15. The dimension of hidden layer in our model is set to 768. Our model is implemented based on PyTorch \cite{Paszke2017AutomaticDI}. In experiments, we use Adamax optimizer for training, with the mini-batch size as 192. For choice of learning rate, we employ the warm-up strategy \cite{Goyal2017AccurateLM}. Specifically, we begin with a learning rate of 5e-4, linearly increasing it at each epoch till it reaches 2e-3. After 11 epochs, the learning rate is decreased by 0.2 for every 2 epochs up to 16 epochs. For transformer encoder, we fix the learning rate as 1e-4. Every linear mapping is regularized by weight normalization and dropout (p = 0.2 except for the classifier with 0.5).

\subsection{Experimental Results}
\noindent
As shown in Table~\ref{tab:vqaval}, our model achieves a superior performance over baseline methods \cite{Anderson2018BottomUpAT,Kim2018BilinearAN,Cadne2019MURELMR,Li2019RelationAwareGA,Gao2019DynamicFW,Gao2019MultiModalityLI}, on VQA 2.0 validation, test-dev and test-std splits. In detail, our method obtains an overall accuracy of 70.51 on test-dev split and 70.71 on test-std split, which surpasses DFAF \cite{Gao2019DynamicFW} and MLIN \cite{Gao2019MultiModalityLI}. Since DFAF and MLIN model dense intra-modal and inter-modal interactions simultaneously, introducing some irrelevant information, such results prove that eliminating irrelevant information can effectively raise the performance. Furthermore, compared with ReGAT\cite{Li2019RelationAwareGA}, our QD-GFN gains improvement of 0.43, 0.24 and 0.13 on validation, test-dev and test-std respectively. Since our model adopts a similar relation encoder as them, above experimental results demonstrate the effectiveness of our proposed GFM and OF. \\
Then, to test the generalization ability of our model, we conduct experiments on VQA-CP v2 dataset, where the distributions of the train and test splits are very different from each other. The results are illustrated in Table~\ref{tab:vqacp}, compared with the previous methods that we have observed on VQA 2.0, QD-GFN surpasses the baseline by a larger margin. 
\begin{table}
    \centering
    \begin{tabular}{l|c|c|c|c}
    \hline
    Model   & All & Y/N & Num & Other  \\
    \hline
    FULL-OF-GFM  & 66.81 & 84.53 & 49.25 & 57.97 \\
    FULL-OF      & 67.07   & 84.9 & 49.26 & 58.22 \\
    FULL-GFM      & 67.29   & 84.91  & 50.19  & 58.41  \\
    FULL  & \textbf{67.61} & \textbf{85.17}  & \textbf{50.38}  & \textbf{58.82} \\
    \hline
    \end{tabular}
    \centering
	\caption{Ablation studies on VQA 2.0 validation split.} 
	\label{tab:vqa_abla}
\end{table}

\begin{figure}[t]
    \begin{center}
    \includegraphics[width=1\linewidth]{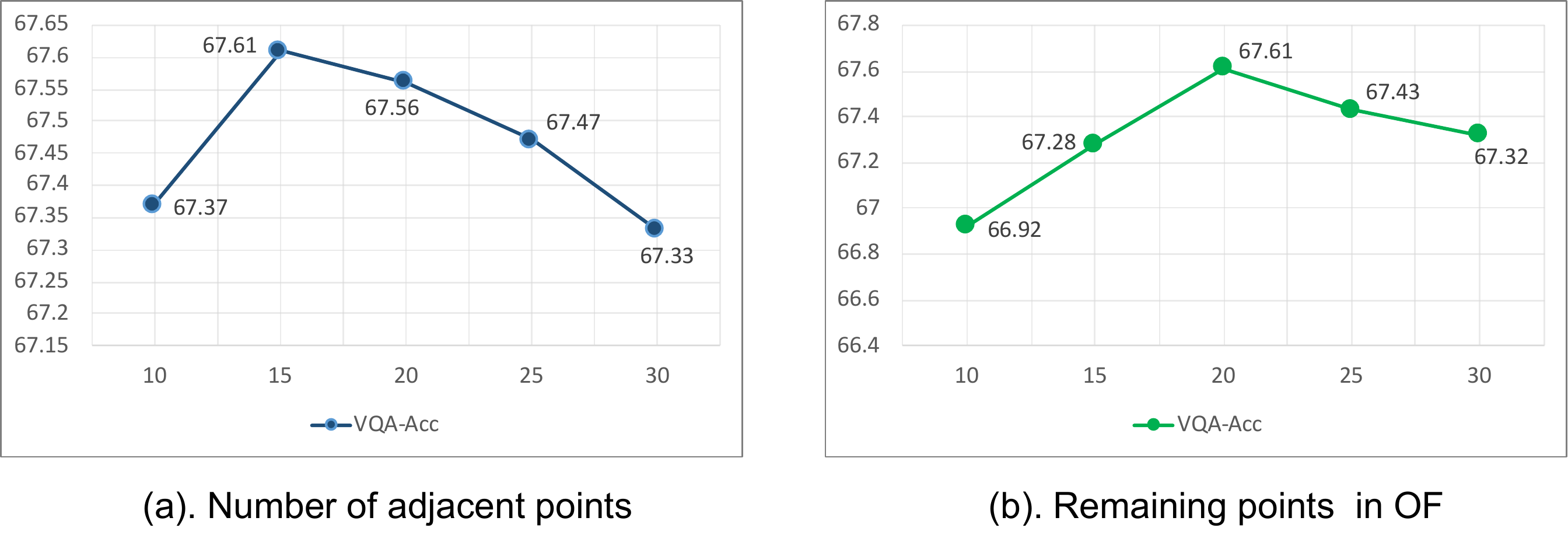}
    \end{center}
    \caption{(a) shows the impact of the number of adjacent nodes and while (b) illustrates the model performance with different number of remained nodes after object filtering}
    \label{fig:vqa_acc}
\end{figure}
\begin{figure}[t]
    \begin{center}
    \includegraphics[width=1\linewidth]{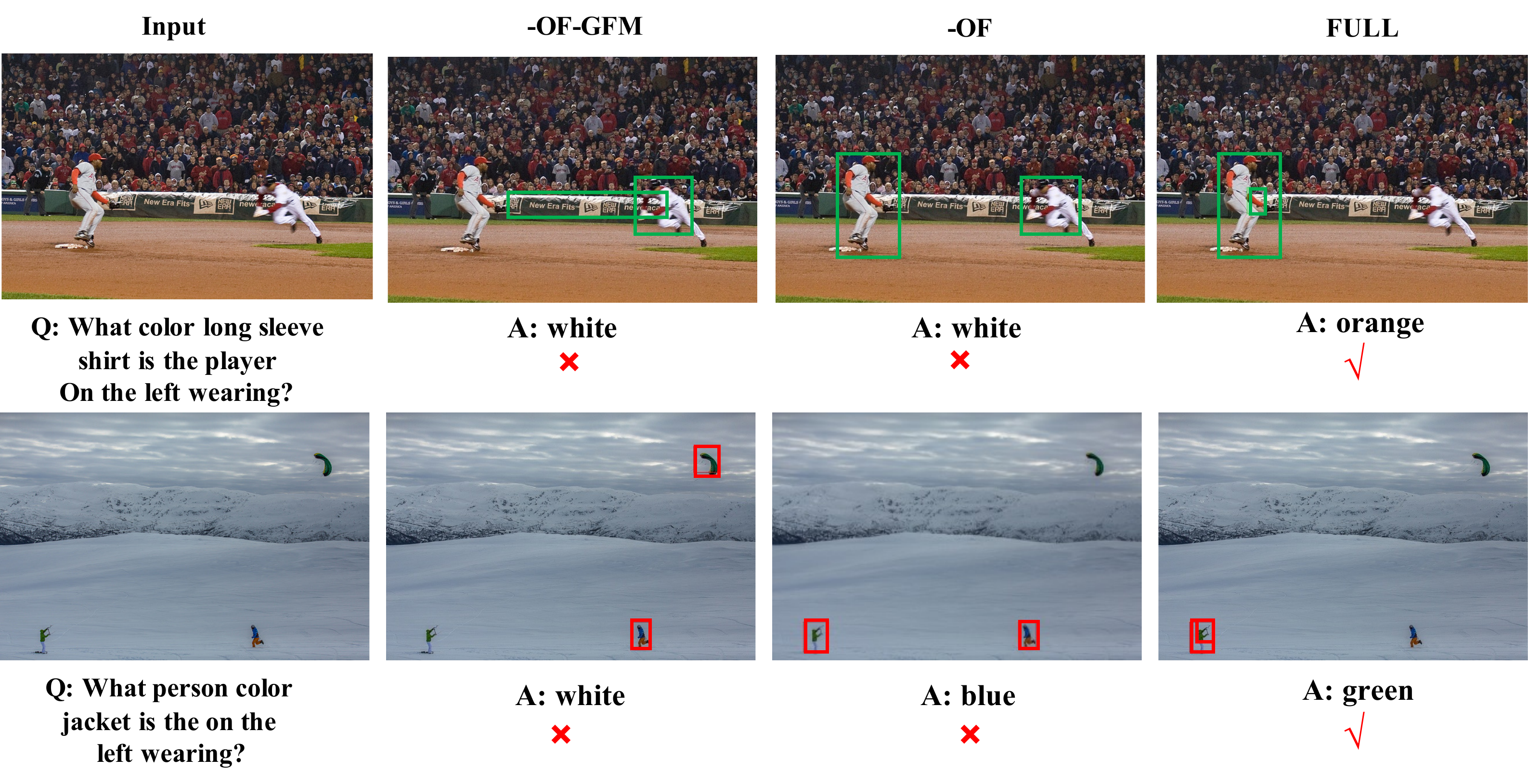}
    \end{center}
    \caption{Visualization of the impact of GFM and OF. The two boxes reserved in each image are the objects with the highest importance coefficient.}
    \label{fig:case_study}
\end{figure}
\subsection{Analysis}
\noindent
 In Table~\ref{tab:vqa_abla}, we conducted several ablation experiments on VQA 2.0 validation split to verify the effectiveness of GFM and OF respectively. Specially, as we can observe from Table~\ref{tab:vqa_abla}, without GFM and OF, the performance decreases significantly to 66.81. Comparison between line 1 and line 2 illustrates a gain of 0.26 for Question-guided Graph Fusion Module, and Object Filtering Module brings an improvement of 0.48 for our model. According to line 4, when GFM and OF are both utilized to give the complete model, it reaches the best performance of 67.61. This result shows that when the OF and GFM modules are combined, they can promote each other and bring greater performance gain.
 
 To further explore the potential factors affecting the performance of the model, fig~\ref{fig:vqa_acc}(a) illustrates the impact of the number of adjacent objects in image relation encoder on the performance of the model, it is obvious to find that setting the number of adjacent objects to 15 is the best choice. Besides, we test the effect of filtering different numbers of objects on the experimental results. As shown in fig~\ref{fig:vqa_acc}(b), when 20 nodes are reserved after filtering, the model achieves the best performance. From the above two cases, we can find that both lacks of information and information redundancy hurt the model. Existing work pays more attention to the introduction of rich information, but ignores the processing of redundant information, which further prominent the value of our work. To more intuitively display the effects of GFM and OF, we visualize the regions of interest in the image. As shown in fig~\ref{fig:case_study}, to comprehensively verify the effectiveness of our proposed QD-GFN, we choose questions containing multiple relationship types as cases. From column 2, when GFM and OF are both eliminated, the model is more easily disturbed and thus attain a wrong answer. By comparing column 2 with column 3, we observe that GFM effectively reduces the wrong attention of the model, and the difference between column 3 and column 4 indicates that object filtering module further eliminates irrelevant objects with strong interference. Column 4 shows that by combining above two modules together, the model completes the filtering of irrelevant information with different granularity, and correctly focuses on question-relevant objects.
\section{Conclusion}
\noindent
We propose Question-Driven Graph Fusion Network(QD-GFN), a novel framework based on image relation encoder for visual question answering, which consists of Question Encoder, Image Relation Encoder, Question-guided Graph Fusion Module(GFM), and Object Filtering Module(OF). Through the GFM and OF modules, our model effectively utilizes information contained in the image and reduces the interference caused by irrelevant information, to achieve competitive results on both VQA 2.0 and VQA-CP v2 datasets. However, the current design of our GFM is still coarse. For future work, we plan to excavate the associated information between text and image more precisely, to better align the information between different modalities.
\subsection*{Acknowledgements} 
\noindent
We first thank anonymous reviewers for their suggestions and comments, and then thank our colleagues for their contributions in providing suggestions for this work. At last, this work was partially supported by the National Natural Science Foundation of China (NSFC62076032) and the Cooperation Project with Beijing SanKuai Technology Co., Ltd. 
\bibliographystyle{IEEEbib}
\bibliography{icme2022template}

\end{document}